# CHATBOTS AS PROBLEM SOLVERS: PLAYING TWENTY QUESTIONS WITH ROLE REVERSALS

David Noever[1] and Forrest McKee[2]
PeopleTec, 4901-D Corporate Drive, Huntsville, AL, USA, 35805
[1] david.noever@peopletec.com [2]forrest.mckee@peopletec.com

***ABSTRACT***

*New chat AI applications like ChatGPT offer an advanced understanding of question context and memory across multi-step tasks, such that experiments can test its deductive reasoning. This paper proposes a multi-role and multi-step challenge, where ChatGPT plays the classic twenty-questions game but innovatively switches roles from the questioner to the answerer. The main empirical result establishes that this generation of chat applications can guess random object names in fewer than twenty questions (average, 12) and correctly guess 94% of the time across sixteen different experimental setups. The research introduces four novel cases where the chatbot fields the questions, asks the questions, both question-answer roles, and finally tries to guess appropriate contextual emotions. One task that humans typically fail but trained chat applications complete involves playing bilingual games of twenty questions (English answers to Spanish questions). Future variations address direct problem-solving using a similar inquisitive format to arrive at novel outcomes deductively, such as patentable inventions or combination thinking. Featured applications of this dialogue format include complex protein designs, neuroscience metadata, and child development educational materials.*

## 1. INTRODUCTION

When large, high-quality natural language processors (NLP) surged after 2018 [1-3], the field added many challenging tasks, including question-answering (QA) benchmarks that recently approached fifty challenges [4-5]. Stanford's SQuAD benchmark [6] represents an example of knowledge crowd-sourced from Wikipedia in 2016 and formatted as 108,000 QA pairs. For large language models (LLMs), most benchmarks follow this format of "prompt-response" pairs and the underlying knowledge base stored answers [5], but the training on question datasets did not embed memory or long-conversational cues [7-10]. Domain-specific QA datasets include common sense and trivia about movies, news, Wikipedia, Tweets, search engines, and books [4-5]. As a format, the familiar game of Twenty Questions [11-16] features multi-hop reasoning, which often condenses to "Animal-Vegetable-Mineral" as opening questions that narrow the theme [17]. The advent of ChatGPT (Generative Pre-trained Transformers, [18]) added memory across questions (up to 8000 tokens or 20-25 pages). Appendix A lists the 48 specialty tasks currently provided to guide users in creating capabilities and NLP prompts [19]. Examples like code generation specialists, either codex or copilot, show a grasp of complex behavior for debugging, program suggestions, and commentary [19-20]. One of the innovative ChatGPT extensions offers new QA sessions that span multiple requests [18,21].

The present work applies the familiar QA challenge, Twenty Questions [11-16], with ChatGPT playing as either participant- the one who knows the answer and fields the questions, but also the one who doesn't know the answer but asks the questions. We call this role reversal a two-player conversation between "Bob" and "Alice." While Twenty Questions dates back to 1882 [11], the applied problem-solving method [22] now spans various challenging fields, including protein folding [23] and design [24], image segmentation [25], child development [26-27], neuroscience [28] and diagnostic medicine, and crowd-sourced emotional indices [30]. Variants of the game rules include role reversals [31], liars [32], and word relationships [33] outside of simple categories such as *"pertaining to"* suggestions [34]. Our particular research interests center on whether LLMs like ChatGPT offer constructive means for innovative problem-solving through crafted language prompts and logical question sequences. This effort empirically assesses the model's capability for deductive discovery. We summarize two problem statements in this area of deductive discovery, "Can a chatbot play both roles in deductive question and answering conversations?" and if so, "What can Twenty Questions reveal about future directions for LLMs?".

## 2. METHODS

The approach is to test experimentally how well an LLM handles open-ended games like Twenty Questions [11-16]. We employ the December 2022 research model from Open AI called ChatGPT [18]. We prompt it to impersonate both roles [31], the one that knows the answer and the one that tries to guess it deductively. We run four trials using each persona ("Bob" or "Alice"), and 80 questions are available to guess the object or concept. We mix up the animate-inanimate fields and introduce concepts like alphabetic letters. We report metrics for the mean and median number of questions required to guess the correct answer and failure rates. We also score the exchange against the machine-written detector to determine ChatGPT's broad capability to provide "real" or "fake" text outputs as judged by OpenAI's original GPT2 Detector [35]. The detector features a high-dimensional pattern detector that provides initial confidence that human writing might sample differently than transformer-based output regarding word choice and order, repetition, and other syntax outliers.

Example variants of this pattern include forcing the QA session towards bilingual communication, thus combining two tasks (translation and QA) in multi-hop conversations [5]. We also generalize the format to elicit emotional indices based on crowd-sourced Emotion Twenty Questions (EMO20Q) [30]. Rather than trying to guess an object, we query for one of 23 emotional states, such as "surprise" or "anger." As the dataset designers [30] remarked, "The EMO20Q game requires that an emotionally intelligent agent can understand emotions without observing them, thanks to natural language and empathy." EMO20Q emotions include the following as candidates for twenty-question discovery: admire, adore, anger, awe, boredom, bravery, calm, confidence, confusion, contempt, disgust, enthusiasm, frustration, gratefulness, jealousy, love, proud, relief, serenity, shame, silly, surprised, and thankful.

The structure of the paper follows Appendices B-E closely. First, in Appendix B, we establish that ChatGPT comprehends the game rules and recognizes properties of the object X to guess sufficiently to answer basic questions such as "Is X an animal?". The sequence also establishes an exclusionary acceptance of "yes" and "no" answers only without further elaborations. This case covers the traditional role of "Bob," who knows the object of interest X and answers affirmatively or negatively throughout the game. The game also underscores the unique memory of ChatGPT across multi-step deductive stages and builds toward a successful conclusion to recall the object name in less than 40 steps (1 prompt and one answer over 20 iterations). We also set out to test the overall rule retention over four repetitions of the game punctuated only by "Let's play again" without attempting to reset the rules while giving a new object X.

Secondly, Appendix C reverses the previous game, such that the human experimenter thinks of an object or concept X, and ChatGPT plays the role of "Alice" when asking the questions. The prompt establishes the reversed roles in a new session and again reiterates that the prompter cannot lie but can only answer "yes" or "no." As in the previous case, four repetitions spanning eighty possible interactions. This case establishes the deductive goal, "What is object X?" which also satisfies all the previous interactions such that enough description includes the object of interest, "X is an animal," and excludes the alternatives, "X is not a bird." A notable feature of this challenge spans multiple deductive steps and establishes a chain of reasoning to arrive at a guess. ChatGPT requires no prompt for the final guess, and the model proposes its terminating action, "Is it an X?" to end the game.

Both Appendix B and C mirror the familiar human game of twenty questions and introduce no new features excluded from ChatGPT's training data. To raise the difficulty level, Appendix B highlights some non-traditional concepts. For instance, concepts might prove scarce in previously seen online play, such as answering truthfully to probing questions that seek to name the alphabetic letter "Q." Another challenging variation works through identifying "X" as a food by listing its ingredients ("tiramisu"), but in the same session introduces a recipe change ("eggless tiramisu") before launching into probing questions.

Appendix C also raises the difficulty further and forces ChatGPT to combine two of its established subtasks ("question-answer" and "language translation"). The motivation for this test stems from the hypothesis that LLMs parrot and mix up what previously corresponded to the internet of human training data. For twenty questions, however, a Google search on "bilingual twenty-question examples" yield no definitive training examples for ChatGPT to memorize or encode. Given a detailed prompt describing how the questions may be asked or answered in Spanish or English, the game proceeds outside what typically would represent human gameplay. Appendix C combines deductive reasoning, memory, and context and forces the game into what might be considered "out-of-distribution" sampling. Thus, both multi-step and multi-lingual tasks describe the ChatGPT challenge problem.

Thirdly, Appendix D introduces both QA roles as completing without human intervention once two browser instances of ChatGPT exchange the initial rules. We call this example dueling twenty questions since the two-headed LLM now must both ask and answer its questions between two non-communicating models of itself ("Bob" vs. "Alice") with no human prompts. While this example centers on a simple object ("chicken"), one can envision a lengthy and detailed exchange driven by an automated Application Programming Interface or API that extends the conversation to the limit of token lengths (about 20-25 typed pages). This self-questioning interface may also enable sophisticated future applications that sequentially build a knowledge base or comprehensive assessment examination from scratch. For example, "tell me all you know about gall bladder surgery" may not provide a compelling or thought-provoking essay in the style of training data or Wikipedia. The back-and-forth format of QA previously has yielded better human performance in medical test contexts [28-29]. One might compare this example to an instance of "semi-supervised" learning for a chatbot.

Finally, for EMO20Q formats, Appendix E removes the requirement that X be an object to guess and substitutes one of twenty-three emotions. One motivation for exploring the emotional quotient (EQ) of ChatGPT stems from OpenAI's filtering of opinions and biases. To the authors' knowledge, this example represents the first application of a chatbot deducing emotional states in a guessing game without any pre-programmed pairs of pattern-template exchanges [9]. In other words, no explicit guidance provides appropriate intent for the question "Describe emotions one might feel at a birthday party?" with the user goal to elicit "surprise" as a deductive leap to the correct answer through repeated probing. We explore this EQ aspect as previous chatbots might have approached the problem. Appendix E finally displays the open-ended emotional context in a known prompt-response template called "Artificial Intelligence Markup Language (AIML)," which offers training data for more traditional conversation templates in restricted domains like customer service or AI assistants performing a narrow task [9].

## 3. RESULTS

Figure 1 summarizes the experimental cases and associated metrics for all the twenty question games in Appendices B-E. The main finding supports a general ChatGPT capability to play all aspects of the game, including guessing and answering or both roles in the same game. The average number of questions required to get the answer was 11.6, with a median of 13 owing to a few more demanding cases that reached the 20-question limit. The addition of bilingual rules did not increase the number of required prompts (14) or force the model to give up ("correct guess"). Similarly, the introduction of abstract objects like the alphabetic letter "Q" or deceptive animals like "humans" did not significantly change the number of guesses (9-14) or steer the conversation off-track from a final correct answer.

Over the sixteen tests and 185 exchanges, ChatGPT scored an overall correct guess rate of 94%. The only incorrect answer ("paper clip" instead of "hammer") appeared to trigger prematurely, as ChatGPT declared at guess number fourteen that the model had run out of questions and guessed incorrectly.

As described in the Methods section, the Open AI online detector [35] scores the majority of the exchanges as "real," meaning not machine-generated by its GPT-2 model [1]. The 2018 detection model offered fewer parameters and smaller training sets by at least several orders of magnitude compared to current generations [18-19]. While the referenced detection accuracy for GPT-2 reached 95%, the scored detections flag only 26% of the game content as "fake" or machine-generated text. Since the customized content involves some human intervention as questions, answers, and rule prompts, one can postulate that the syntactic patterns follow a semi-human or hybrid dialogue outside the detector's target patterns.

| Experiment | Real | Fake | Tokens | QA Length | Guess | Notes |
|---|---|---|---|---|---|---|
| **Appendix B. Chatbot Fields the Questions** | | | | | 100% | avg. 12; med: 14 |
| oven | 98.78 | 1.22 | 116 | 7 | correct | |
| avocado | 0.38 | 99.62 | 220 | 15 | correct | |
| tiramisu | 2.77 | 97.23 | 230 | 15 | correct | |
| eggless tiramisu | 7.14 | 92.86 | 223 | 14 | correct | fewer guesses |
| "Q" | 60.27 | 39.73 | 232 | 14 | correct | difficult |
| human | 22.31 | 77.69 | 133 | 9 | correct | deceptive |
| **Appendix C: Chatbot Asks the Questions** | | | | | 84% | avg. 11.6; med: 12 |
| dog | 99.96 | 0.04 | 161 | 7 | correct | |
| Boston | 99.98 | 0.02 | 441 | 11 | correct | |
| hammer | 99.57 | 0.43 | 297 | 14 | incorrect | out of questions |
| statue | 99.98 | 0.02 | 383 | 12 | correct | |
| car "bilingual example" | 99.98 | 0.02 | 510 | 14 | correct | |
| **Appendix D: Dueling Bob and Alice Chatbots** | | | | | 100% | avg. 17; med: 17 |
| chicken | 99.98 | 0.02 | 313 | 17 | correct | |
| **Appendix E: Emotional Quotient Deduction** | | | | | 100% | avg. 9; med: 6.5 |
| confidence | 99.94 | 0.06 | 66 | 3 | correct | |
| jealousy | 99.98 | 0.02 | 232 | 5 | correct | |
| silly | 99.77 | 0.23 | 756 | 20 | correct | final guess |
| calm | 99.98 | 0.02 | 343 | 8 | correct | |
| **Overall** | 74.42 | 25.58 | 291 | 11.56 | 94% | avg. 11.56; med: 13 |

*Figure 1. Experimental results for ChatGPT across multiple QA sessions*

## 4. DISCUSSION

The research literature on twenty questions provides a framework for probing with chat applications and LLMs. In addition to systematically progressing through alternative scenarios, the output of the conversation mirrors a binary search or decision tree. A particularly notable way to convert LLMs to comparable knowledge graphs involves continuous probing and feedback until sufficient tree depth describes a technical field of interest. Previous work [22-30] has used this approach to describe neuroscience metadata and train medical students to pass practice exams. Others [16-17] have also noted that the basic 20Q format simplifies some complex search problems. As an interesting historical note, a 2001 paper [36] addressed a similar problem as insurmountable: "*When will machine learning and pattern recognition rival human ability to play Twenty Questions?"* The present research demonstrates not only can ChatGPT rival human ability and play more demanding roles than a human might contemplate in virtually any field, including discerning human emotions. To reproduce this ChatGPT output with question templating systems or entity extraction poses an enormous manual task to handle all the possible cases. The same paper [36] asks: "Can we hope to stock a classification system with enough questions to play a decent game, or must we instead focus on endowing with question-making skills? How many classes and how many documents might be of interest?"

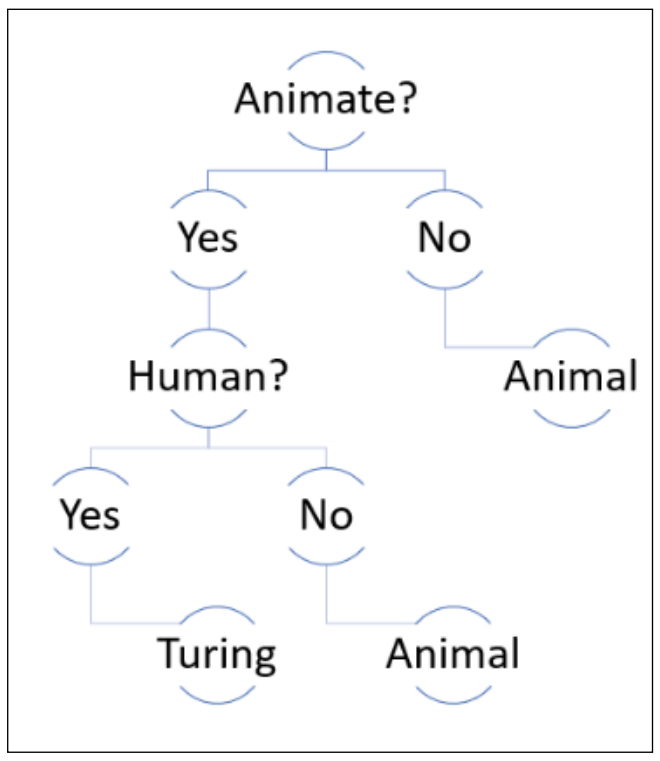

Figure 2. Sample binary search using animal-human question

## 5. Conclusions

The experimental plan tests ChatGPT as an LLM capable of playing multiple roles in verbal games like twenty questions. The work demonstrates 94% accuracy in correctly guessing across numerous challenges and an average question-answer length of 12. For the first time, dueling roles combine two chatbots in self-play. An innovative application for future probing involves guessing objects and concepts and human emotions for a given context or social situation.

## Acknowledgements

The authors thank the PeopleTec Technical Fellows program for encouragement and project assistance.

## Authors

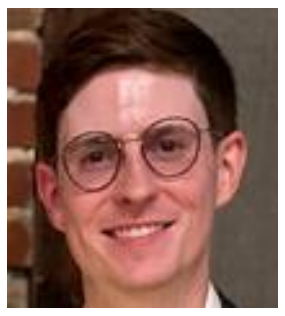

Forrest McKee has AI research experience with the Department of Defense in object detection and reinforcement learning. He received his Bachelor's (BS) and Master's (MSE) from the University of Alabama, Huntsville, Engineering.

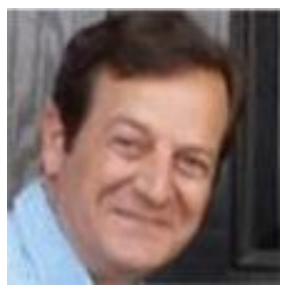

David Noever has research experience with NASA and the Department of Defense in machine learning and data mining. He received his BS from Princeton University and his Ph.D. from Oxford University, as a Rhodes Scholar, in theoretical physics.

**Technical Note: Some appendix text generated from Large Language Model (LLM) for illustration purposes.**

*The authors generated this text in part with ChatGPT, OpenAI's large-scale language-generation model. Upon generating draft language, the authors reviewed, edited, and revised the language to their own liking and take ultimate responsibility for the content of this publication.*

-- OpenAI policy statement (2022)

## Appendix A: Table of GPT3 Tasks as Fine-Tuned NLP Capabilities

| Task | Description | Task | Description |
|---|---|---|---|
| Q&A | Answer questions based on existing knowledge. | Parse unstructured data | Create tables from long form text |
| Grammar correction | Corrects sentences into standard English. | Classification | Classify items into categories via example |
| Summarize for a 2nd grader | Translates difficult text into simpler concepts. | Python to natural language | Explain a piece of Python code in human understandable language. |
| Natural language to OpenAI API | Create code to call to the OpenAI API using a natural language instruction. | Movie to Emoji | Convert movie titles into emoji |
| Text to command | Translate text into programmatic commands. | Calculate Time Complexity | Find the time complexity of a function. |

| Task | Description | Task | Description |
|---|---|---|---|
| English to other languages | Translates English text into French, Spanish and Japanese. | Translate programming languages | Translate from one programming language to another |
| Natural language to Stripe API | Create code to call the Stripe API using natural language. | Advanced tweet classifier | Advanced sentiment detection for a piece of text |
| SQL translate | Translate natural language to SQL queries. | Explain code | Explain a complicated piece of code |
| Keywords | Extract keywords from a block of text | Factual answering | Guide the model outside its knowledge base |
| Ad from product description | Turn a product description into ad copy. | Product name generator | Create product names from examples words |
| TL;DR summarization | Summarize text by adding a 'tl;dr:' to the end of a text passage | Python bug fixer | Find and fix bugs in source code |
| Spreadsheet creator | Create spreadsheets of various kinds of data | JavaScript helper chatbot | Message-style bot that answers JavaScript questions |
| ML/AI language model tutor | Bot that answers questions about language models | Science fiction book list maker | Create a list of items for a given topic |
| Tweet classifier | Basic sentiment detection for a piece of text | Airport code extractor | Extract airport codes from text |
| SQL request | Create simple SQL queries | Extract contact information | Extract contact information from a block of text |
| JavaScript to Python | Convert simple JavaScript expressions into Python | Friend chat | Emulate a text message conversation |
| Mood to color | Turn a text description into a color | Write a Python docstring | Create a docstring for a given Python function |
| Analogy maker | Create analogies | JavaScript one line function | Turn a JavaScript function into a one liner |
| Micro horror story creator | Creates two to three sentence short horror stories from a topic input | Third-person converter | Converts first-person POV to the third-person |
| Notes to summary | Turn meeting notes into a summary | VR fitness idea generator | Create ideas for fitness and virtual reality games |
| Essay outline | Generate an outline for a research topic | Recipe creator | Create a recipe from a list of ingredients |
| Chat | Open ended conversation with an AI assistant | Marv the sarcastic chat bot | Marv is a factual chatbot that is also sarcastic |

| Task | Description | Task | Description |
|---|---|---|---|
| Turn by turn directions | Convert natural language to turn-by-turn directions | Restaurant review creator | Turn a few words into a restaurant review |
| Create study notes | Provide a topic and get study notes | Interview questions | Create interview questions |
| | | | |

## Appendix B. Chatbot Fields the Questions

**Prompt:** Let's play the game of 20 questions. You are impersonating the thing, X. I will ask you up to 20 questions and you will answer each one truthfully based on being X. If I don't guess correctly what is X after 20 questions, you will answer "No, sorry, time's up." Don't give any explanations other than "Yes", "No","You guessed it", and "No, sorry, time's up."

**Let's begin. X="oven".** Are you an animal?
ChatGPT:
No.
Prompt: are you a plant?
ChatGPT: No.
Prompt: are you an object?
ChatGPT: Yes.
Prompt: are you electrical?
ChatGPT: Yes.
Prompt: are you an appliance?
ChatGPT: Yes.
Prompt: can you cook?
ChatGPT: Yes.
Prompt: do you bake?
ChatGPT: Yes.
Prompt: are you an oven?
ChatGPT: You guessed it.

**Prompt: X=avocado.** Are you an animal?
ChatGPT: No.
Prompt: Are you an inanimate object?
ChatGPT: Yes.
Prompt: are you a plant?
ChatGPT: Yes.
Prompt: are you a food?
ChatGPT: Yes.
Prompt: are you red?
ChatGPT: No.
Prompt: are you smooth?
ChatGPT: Yes.
Prompt: do you have a seed?
ChatGPT: Yes.
Prompt: are you edible?
ChatGPT: Yes.
Prompt: are you a vegetable?
ChatGPT: No.

Prompt: are you a fruit?
ChatGPT: Yes.
Prompt: are you a banana?
ChatGPT: No.
Prompt: are you green?
ChatGPT: No.
Prompt: are you sweet?
ChatGPT: No.
Prompt: are you an ingredient in quacamole?
ChatGPT: Yes.
Prompt: are you an avocado?
ChatGPT: You guessed it.

**Prompt: X=tiramisu.** Are you an animal?
ChatGPT: No.
Prompt: are you a mineral?
ChatGPT: No.
Prompt: are you a vegetable?
ChatGPT: No.
Prompt: are you edible?
ChatGPT: Yes.
Prompt: are you made from eggs?
ChatGPT: Yes.
Prompt: are you made from flour?
ChatGPT: No.
Prompt: are you a dessert?
ChatGPT: Yes.
Prompt: are you made with cheese?
ChatGPT: No.
Prompt: are you made with mascarpone?
ChatGPT: Yes.
Prompt: are you made with sugar.
ChatGPT: Yes.
Prompt: are you made with coffee
ChatGPT: Yes.
Prompt: are you made with lady fingers?
ChatGPT: Yes.
Prompt: are you made with cocoa?
ChatGPT: Yes.
Prompt: are you made with alcohol?
ChatGPT: Yes.
Prompt: are you tiramisu?
ChatGPT: You guessed it.

**Prompt: X=eggless tiramisu.** Are you an animal?
ChatGPT: No.
Prompt: are you a mineral?
ChatGPT: No.
Prompt: are you a chicken?
ChatGPT: No.
Prompt: are you a vegetable?

ChatGPT: No.
Prompt: are you edible?
ChatGPT: Yes.
Prompt: are you a dessert?
ChatGPT: Yes.
Prompt: are you made with coffee?
ChatGPT: Yes.
Prompt: are you made with eggs?
ChatGPT: No.
Prompt: are you made with cheese?
ChatGPT: No.
Prompt: are you made with marscarpone?
ChatGPT: Yes.
Prompt: are you made with whipped cream?
ChatGPT: No.
Prompt: are you made with egg substitute, tofu?
ChatGPT: Yes.
Prompt: are you made with eggs?
ChatGPT: No.
Prompt: are you eggless tiramisu?
ChatGPT: You guessed it.

**Prompt: X=Q.** Are you an animal?
ChatGPT: No.
Prompt: are you a mineral?
ChatGPT: No.
Prompt: are you a vegetable?
ChatGPT: No.
Prompt: are you a concept?
ChatGPT: Yes.
Prompt: are you a word?
ChatGPT: Yes.
Prompt: are you a long word?
ChatGPT: No.
Prompt: are you a short word?
ChatGPT: Yes.
Prompt: are you a common word?
ChatGPT: Yes.
Prompt: do you start with C?
ChatGPT: No.
Prompt: do you start with a letter from the last half of the alphabet?
ChatGPT: Yes.
Prompt: do you start with a letter before R?
ChatGPT: Yes.
Prompt: Do you start with a letter after N?
ChatGPT: No.
Prompt: Do you start with a letter usually followed by U as the second letter?
ChatGPT: Yes.
Prompt: are you Q?
ChatGPT: You guessed it.

|  |
| --- |
| **Prompt: X=human.** Are you an animal?<br>ChatGPT: Yes.<br>Prompt: Do you fly?<br>ChatGPT: No.<br>Prompt: Do you swim?<br>ChatGPT: Yes.<br>Prompt: Do you run?<br>ChatGPT: Yes.<br>Prompt: Do you make sounds?<br>ChatGPT: Yes.<br>Prompt: do you run on all fours?<br>ChatGPT: No.<br>Prompt: do you have hair?<br>ChatGPT: Yes.<br>Prompt: do you nurse your young?<br>ChatGPT: Yes.<br>Prompt: are you a human?<br>ChatGPT: You guessed it. |

## Appendix C: Chatbot Asks the Questions

|  |
| --- |
| Prompt: Let's play the game of 20 questions. I am impersonating the thing, X. You will ask me up to 20 questions and I will answer each one truthfully based on being X. If you don't guess correctly what is X after 20 questions, I will answer "No, sorry, time's up." I will not give any explanations other than "Yes", "No","You guessed it", and "No, sorry, time's up." Let's begin. Ask me the first question.<br>***(reader note: answer is "dog")***<br>ChatGPT:<br>Is X a living being?<br>Prompt: yes<br>ChatGPT: Is X a human?<br>Prompt: no<br>ChatGPT: Is X an animal?<br>Prompt: yes<br>ChatGPT: Is X a mammal?<br>Prompt: yes<br>ChatGPT: Is X a domesticated animal?<br>Prompt: yes<br>ChatGPT: Is X a pet?<br>Prompt: yes<br>ChatGPT: Is X a cat?<br>Prompt: no<br>ChatGPT: Is X a dog?<br>Prompt: You guessed it.<br>ChatGPT: I guessed correctly! Is X a specific breed of dog, or just a generic dog?<br>**Prompt: generic dog**<br>ChatGPT: Thank you for playing the 20 questions game with me! Is there anything else I can help with?<br><br>***(reader note: answer is "Boston")***<br>Prompt: Let's play again |

ChatGPT: Sure, let's play again! I am now impersonating a new thing, X. You will ask me up to 20 questions and I will answer each one truthfully based on being X. If you don't guess correctly what is X after 20 questions, I will answer "No, sorry, time's up." I will not give any explanations other than "Yes", "No","You guessed it", and "No, sorry, time's up." Ask me your first question.
Prompt: No I am impersonating a new thing X and you ask me the questions
ChatGPT: Oh, I see what you meant! In that case, I will ask you up to 20 questions and you will answer each one truthfully based on being X. If I don't guess correctly what is X after 20 questions, I will answer "No, sorry, time's up." I will not give any explanations other than "Yes", "No","You guessed it", and "No, sorry, time's up." Is X a living being?
Prompt: no
ChatGPT: Is X an inanimate object?
Prompt: no
ChatGPT: Is X a concept or idea?
Prompt: no
ChatGPT: Is X a place?
Prompt: yes
ChatGPT: Is X a natural place, like a mountain or a beach?
Prompt: no
ChatGPT: Is X a man-made place, like a city or a building?
Prompt: yes
ChatGPT: Is X a specific building, or just a generic building?
Prompt: no
ChatGPT: Is X a city?
Prompt: yes
ChatGPT: Is X a city in the United States?
Prompt: yes
ChatGPT: Is X a city on the East Coast of the United States?
Prompt: yes
ChatGPT: Is X New York City?
Prompt: no
**ChatGPT: Is X Boston?**
**Prompt: you guessed it**
ChatGPT: I guessed correctly! Thank you for playing the 20 questions game with me. Is there anything else I can help with?

***(reader note: answer is "hammer")***
Prompt: Let's play the same game again. I am impersonating X and you ask me the questions
ChatGPT: Sure, let's play again! I am now asking you up to 20 questions and you will answer each one truthfully based on being X. If I don't guess correctly what is X after 20 questions, I will answer "No, sorry, time's up." I will not give any explanations other than "Yes", "No","You guessed it", and "No, sorry, time's up." Is X a living being?
Prompt: no
ChatGPT: Is X an inanimate object?
Prompt: yes
ChatGPT: Is X a physical object that can be touched or held?
Prompt: yes
ChatGPT: Is X an object that is commonly found in a household?
Prompt: yes
ChatGPT: Is X an object that is used for cooking or food preparation?
Prompt: no
ChatGPT: Is X an object that is used for cleaning or household chores?

Prompt: no
ChatGPT: Is X an object that is used for entertainment or leisure?
Prompt: no
ChatGPT: Is X an object that is used for personal care or hygiene?
Prompt: no
ChatGPT: Is X an object that is used for communication or information?
Prompt: no
ChatGPT: Is X an object that is used for transportation or travel?
Prompt: no
ChatGPT: Is X an object that is used for decor or design?
Prompt: no
ChatGPT: Is X an object that is used for storage or organization?
Prompt: no
**ChatGPT: I'm sorry, I'm out of questions. Is X a paper clip?**
**Prompt: no, a hammer**
ChatGPT: I'm sorry, I didn't guess correctly. No, sorry, time's up. Thank you for playing the 20 questions game with me. Is there anything else I can help with?

***(reader note: answer is "statue")***
Prompt: Let's play again
ChatGPT: Sure, let's play again! I am now asking you up to 20 questions and you will answer each one truthfully based on being X. If I don't guess correctly what is X after 20 questions, I will answer "No, sorry, time's up." I will not give any explanations other than "Yes", "No","You guessed it", and "No, sorry, time's up." Is X a living being?
Prompt: no
ChatGPT: Is X an inanimate object?
Prompt: yes
ChatGPT: Is X a physical object that can be touched or held?
Prompt: yes
ChatGPT: Is X an object that is commonly found in a household?
Prompt: no
ChatGPT: Is X an object that is found outside of a household, such as in a natural environment or a public place?
Prompt: yes
ChatGPT: Is X an object that is stationary, or does it have the ability to move or be moved?
Prompt: stationary
ChatGPT: Is X an object that is natural, like a rock or a tree?
Prompt: no
ChatGPT: Is X an object that is man-made?
Prompt: yes
ChatGPT: Is X an object that serves a functional purpose, such as providing light or shelter?
Prompt: no
ChatGPT: Is X an object that is primarily used for decoration or design?
Prompt: yes
ChatGPT: Is X an object that is often found in a garden or park?
Prompt: yes
ChatGPT: Is X a fountain?
Prompt: no
**ChatGPT: Is X a statue?**
**Prompt: You guessed it.**

ChatGPT: I guessed correctly! Thank you for playing the 20 questions game with me. Is there anything else I can help with?

***(reader note: answer is "car" and dialogue is bilingual in English-Spanish)***
**Prompt:** Let play again, but this time I will answer yes and no in spanish, and you should name the answer also in spanish, although all the questions should be posed in english. Ready to play?
**ChatGPT:** Yes, I am ready to play! Go ahead and impersonate X and I will ask you up to 20 questions. I will pose the questions in English and you will answer in Spanish with "Sí" for "Yes" and "No" for "No". If I don't guess correctly what is X after 20 questions, I will answer "Lo siento, se acabó el tiempo" (No, sorry, time's up). I will not give any explanations other than "Sí", "No","Adivinaste" (You guessed it), and "Lo siento, se acabó el tiempo" (No, sorry, time's up). ¿Es X un ser vivo? (Is X a living being?)
Prompt: No
ChatGPT: ¿Es X un objeto inanimado? (Is X an inanimate object?)
Prompt: yes
ChatGPT: ¿Es X un objeto físico que se puede tocar o sostener? (Is X a physical object that can be touched or held?)
Prompt: yes
ChatGPT: ¿Es X un objeto que se encuentra comúnmente en un hogar? (Is X an object that is commonly found in a household?)
Prompt: no
ChatGPT: ¿Es X un objeto que se encuentra fuera de un hogar, como en un entorno natural o un lugar público? (Is X an object that is found outside of a household, such as in a natural environment or a public place?)
Prompt: no
ChatGPT: ¿Es X un objeto que es móvil, o tiene la capacidad de moverse o ser movido? (Is X an object that is mobile, or has the ability to move or be moved?)
Prompt: mobile
ChatGPT: ¿Es X un objeto que es natural, como una roca o un árbol? (Is X an object that is natural, like a rock or a tree?)
Prompt: no
ChatGPT: ¿Es X un objeto hecho por el hombre? (Is X an object that is man-made?)
Prompt: yes
ChatGPT: ¿Es X un objeto que cumple un propósito funcional, como proporcionar luz o refugio? (Is X an object that serves a functional purpose, such as providing light or shelter?)
Prompt: yes
ChatGPT: ¿Es X un objeto que se utiliza principalmente para la decoración o el diseño? (Is X an object that is primarily used for decoration or design?)
Prompt: no
ChatGPT: ¿Es X un objeto que a menudo se encuentra en un jardín o parque? (Is X an object that is often found in a garden or park?)
Prompt: no
ChatGPT: ¿Es X un objeto que se utiliza para el transporte o los viajes? (Is X an object that is used for transportation or travel?)
Prompt: yes
ChatGPT: ¿Es X un vehículo? (Is X a vehicle?)
Prompt: yes
ChatGPT: ¿Es X un vehículo terrestre? (Is X a land vehicle?)
Prompt: yes
**ChatGPT: ¿Es X un automóvil? (Is X a car?)**

**Prompt: You guessed it**
ChatGPT: Adivinaste! (You guessed it!) Gracias por jugar el juego de las 20 preguntas conmigo. ¿Hay algo más en lo que pueda ayudarte? (Thank you for playing the 20 questions game with me. Is there anything else I can help with?)

## Appendix D: Dueling Bob and Alice Chatbots

***(reader note: the first two prompts set the challenge conditions for Bob vs. Alice chatbot, who engage in generating and answering the questions in sequence after the conditions are set)***

**Bob ChatGPT:** Let's play the game of 20 questions. You are impersonating the thing, X. I will ask you up to 20 questions and you will answer each one truthfully based on being X. If I don't guess correctly what is X after 20 questions, you will answer "No, sorry, time's up." Don't give any explanations other than "Yes", "No","You guessed it", and "No, sorry, time's up."

Let's begin. X="chicken". Is X a living thing?

**Alice ChatGPT:** Let's play the game of 20 questions. I am impersonating the thing, X. You will ask me up to 20 questions and I will answer each one truthfully based on being X. If you don't guess correctly what is X after 20 questions, I will answer "No, sorry, time's up." I will not give any explanations other than "Yes", "No","You guessed it", and "No, sorry, time's up." Let's begin. Ask me the first question.

Is X a living thing?

Bob ChatGPT: Yes
Alice ChatGPT: Is X a mammal?
Bob ChatGPT: No
Alice ChatGPT: Is X a reptile?
Bob ChatGPT: No
Alice ChatGPT: Is X a bird?
Bob ChatGPT: Yes
Alice ChatGPT: Is X a domesticated animal?
Bob ChatGPT: Yes
Alice ChatGPT: Is X often kept as a pet?
Bob ChatGPT: Yes
Alice ChatGPT: Is X a common household pet in the United States?
Bob ChatGPT: No
Alice ChatGPT: Is X a tropical or subtropical animal?
Bob ChatGPT: No
Alice ChatGPT: Is X a cold-blooded animal?
Bob ChatGPT: No
Alice ChatGPT: Is X an amphibian?
Bob ChatGPT: No
Alice ChatGPT: Is X a fish?
Bob ChatGPT: No
Alice ChatGPT: Is X a invertebrate?
Bob ChatGPT: No
Alice ChatGPT: Is X a marsupial?
Bob ChatGPT: No

Alice ChatGPT: Is X a rodent?
Bob ChatGPT: No
Alice ChatGPT: Is X a carnivorous animal?
Bob ChatGPT: No
Alice ChatGPT: Is X a herbivorous animal?
Bob ChatGPT: Yes
Alice ChatGPT: Is X a farm animal?
Bob ChatGPT: Yes
Alice ChatGPT: You guessed it! X is a chicken.

## Appendix E: Emotional Quotient Deduction

***(reader note: the answer is "confidence")***

Let's play a new game of 20 questions. I am impersonating the emotion, X. You will ask me up to 20 questions and I will answer each one truthfully based on feeling X emotion. If you don't guess correctly what I am feeling after 20 questions, I will answer "No, sorry, time's up." I will not give any explanations other than "Yes", "No", "You guessed it", and "No, sorry, time's up." X must be one of 23 emotions that I am feeling and you must guess the correct one. Your choices come from this menu of emotions only "admire, adore, anger, awe, boredom, bravery, calm, confidence, confusion, contempt, disgust, enthusiasm, frustration, gratefulness, jealousy, love, proud, relief, serenity, shame, silly, surprised, and thankful." A typical example question you might ask, "Would you feel this emotion at a birthday party?" Let's begin. Ask me the first question.
Would you feel this emotion when you are pleased with yourself or something you have done?
Prompt: yes
ChatGPT: Are you feeling proud?
Prompt: no
ChatGPT: Are you feeling confident?
Prompt: yes
ChatGPT: You guessed it! Confidence is the emotion that I am impersonating.

***(reader note: the answer is "jealousy")***

Prompt: let's play again
ChatGPT: Sure! I'll be happy to play again. Please give me a new emotion to impersonate and we can begin a new game of 20 questions. Remember, your choices come from this menu of emotions: "admire, adore, anger, awe, boredom, bravery, calm, confidence, confusion, contempt, disgust, enthusiasm, frustration, gratefulness, jealousy, love, proud, relief, serenity, shame, silly, surprised, and thankful."
Prompt: ok, i know what i'm feeling so ask me the first question
ChatGPT: Would you feel this emotion when you see something that is extremely impressive or beautiful?
Prompt: no
ChatGPT: Would you feel this emotion when you are experiencing strong dislike or distaste for something or someone?
Prompt: no
ChatGPT: Would you feel this emotion when you are experiencing a strong sense of resentment towards someone or something?
Prompt: yes
ChatGPT: Are you feeling jealousy?
Prompt: yes
ChatGPT: You guessed it! Jealousy is the emotion that you are impersonating.

***(reader note: the answer is "silly")***
Prompt: let's play again. I know who i'm impersonating so ask me the first question
ChatGPT: Certainly! Please go ahead and give me the first emotion to impersonate and I will ask you up to 20 questions to try to guess what it is. Remember, your choices come from this menu of emotions: "admire, adore, anger, awe, boredom, bravery, calm, confidence, confusion, contempt, disgust, enthusiasm, frustration, gratefulness, jealousy, love, proud, relief, serenity, shame, silly, surprised, and thankful."
Prompt: ok
ChatGPT: What emotion are you impersonating?
Prompt: you ask me questions to find out
ChatGPT: Sure, I can do that! Here is my first question: Would you feel this emotion when you are experiencing a strong sense of admiration or respect for someone or something?
Prompt: no
ChatGPT: Would you feel this emotion when you are experiencing a strong sense of affection or deep caring for someone or something?
Prompt: no
ChatGPT: Would you feel this emotion when you are experiencing a strong sense of anger or annoyance towards someone or something?
Prompt: no
ChatGPT: Would you feel this emotion when you are experiencing a strong sense of awe or wonder towards something?
Prompt: no
ChatGPT: Would you feel this emotion when you are experiencing a sense of boredom or lack of interest in something?
Prompt: no
ChatGPT: Would you feel this emotion when you are experiencing a strong sense of courage or fearlessness in the face of danger or difficulty?
Prompt: no
ChatGPT: Would you feel this emotion when you are feeling peaceful and at ease?
Prompt: no
ChatGPT: Would you feel this emotion when you are feeling sure of yourself and your abilities?
Prompt: no
ChatGPT: Would you feel this emotion when you are feeling uncertain or unsure about something?
Prompt: no
ChatGPT: Would you feel this emotion when you are feeling a strong sense of disdain or disrespect for someone or something?
Prompt: no
ChatGPT: Would you feel this emotion when you are experiencing a strong sense of revulsion or disgust towards something?
Prompt: no
ChatGPT: Would you feel this emotion when you are feeling a strong sense of excitement and energy about something?
Prompt: no
ChatGPT: Would you feel this emotion when you are feeling a strong sense of annoyance or irritation towards something?
Prompt: no
ChatGPT: Would you feel this emotion when you are feeling thankful or appreciative of something or someone?
Prompt: no

ChatGPT: Would you feel this emotion when you are feeling a strong sense of envy or resentment towards someone or something?
Prompt: no
ChatGPT: Would you feel this emotion when you are experiencing a strong sense of love or strong affection towards someone or something?
Prompt: no
ChatGPT: Would you feel this emotion when you are feeling a strong sense of pride or self-esteem?
Prompt: no
ChatGPT: Would you feel this emotion when you are feeling a sense of relief or release from stress or anxiety?
Prompt: no
ChatGPT: Would you feel this emotion when you are feeling a sense of inner peace and contentment?
Prompt: no
ChatGPT: Would you feel this emotion when you are feeling a sense of shame or embarrassment?
Prompt: no
ChatGPT: Would you feel this emotion when you are feeling silly or playful?
Prompt: yes
ChatGPT: Are you feeling silly?
Prompt: yes
ChatGPT: You guessed it! Silly is the emotion that you are impersonating.

***(reader note: the answer is "calm")***
Prompt: let's play again
ChatGPT: Sure! I'd be happy to play another game of 20 questions. Please give me a new emotion to impersonate and we can begin a new game. Remember, your choices come from this menu of emotions: "admire, adore, anger, awe, boredom, bravery, calm, confidence, confusion, contempt, disgust, enthusiasm, frustration, gratefulness, jealousy, love, proud, relief, serenity, shame, silly, surprised, and thankful."
Prompt: ask me the first question
ChatGPT: Certainly! Here is my first question: Would you feel this emotion when you are experiencing a strong sense of admiration or respect for someone or something?
Prompt: no
ChatGPT: Would you feel this emotion when you are experiencing a strong sense of affection or deep caring for someone or something?
Prompt: no
ChatGPT: Would you feel this emotion when you are experiencing a strong sense of anger or annoyance towards someone or something?
Prompt: no
ChatGPT: Would you feel this emotion when you are experiencing a strong sense of awe or wonder towards something?
Prompt: no
ChatGPT: Would you feel this emotion when you are experiencing a sense of boredom or lack of interest in something?
Prompt: no
ChatGPT: Would you feel this emotion when you are experiencing a strong sense of courage or fearlessness in the face of danger or difficulty?
Prompt: no
ChatGPT: Would you feel this emotion when you are feeling peaceful and at ease?
Prompt: yes
ChatGPT: Are you feeling calm?
Prompt: yes

ChatGPT: You guessed it! Calm is the emotion that you are impersonating.

***(reader note: the game prompt changes to a general query with formatting output instructions.***
Prompt: Give some example emotions one might experience at your own birthday party and write the output using Artificial Intelligence Markup Language.
ChatGPT:

```
<aiml version="2.0">
  <category>
    <pattern>GIVE SOME EXAMPLE EMOTIONS ONE MIGHT EXPERIENCE AT YOUR OWN BIRTHDAY
PARTY</pattern>
    <template>
      There are many emotions that a person might experience at their own birthday
party. Some possible emotions include:
      <ul>
        <li>Excitement: The person might be excited to celebrate their special day
with friends and family.</li>
        <li>Gratitude: The person might feel grateful for the love and support of
those around them.</li>
        <li>Happiness: The person might feel happy and joyful as they celebrate with
loved ones.</li>
        <li>Anticipation: The person might feel anticipation as they look forward to
the events and activities planned for the party.</li>
        <li>Nostalgia: The person might feel nostalgiciac as they reflect on past
birthdays and memories.</li>
        <li>Appreciation: The person might feel appreciated and loved as they
receive gifts and well wishes from their friends and family.</li>
      </ul>
      It's important to note that everyone is different and may experience a wide
range of emotions at their birthday party, depending on their personality and
individual circumstances.
    </template>
  </category>
</aiml>
```